# A Faster Patch Ordering Method for Image Denoising

Badre Munir

*Abstract*— Among the patch-based image denoising processing methods, smooth ordering of local patches (patch ordering) has been shown to give state-of-art results. For image denoising the patch ordering method forms two large TSPs (Traveling Salesman Problem) comprised of nodes in N-dimensional space. Ten approximate solutions of the two large TSPs are then used in a filtering process to form the reconstructed image. Use of large TSPs makes patch ordering a computationally intensive method. A modified patch ordering method for image denoising is proposed. In the proposed method, several smaller-sized TSPs are formed and the filtering process varied to work with solutions of these smaller TSPs. In terms of PSNR, denoising results of the proposed method differed by 0.032 dB to 0.016 dB on average. In original method, solving TSPs was observed to consume 85% of execution time. In proposed method, the time for solving TSPs can be reduced to half of the time required in original method. The proposed method can denoise images in 40% less time.

## I. INTRODUCTION

SEVERAL image denoising methods employing local image patches have been developed lately [1]–[5]. Among these methods, smooth ordering of local patches (patch ordering) [6] has been shown to give state-of-art results for image denoising giving comparable performance to the BM3D algorithm [7]. However, patch ordering method has been reported as a computationally intensive because it employs solutions of large Traveling Salesman Problems (TSPs).

For image denoising, the patch ordering method forms two large TSPs with nodes in $\mathbb{R}^n$; depending on noise intensity $n$ ranges from 25 to 64. Ten approximate solutions of both TSPs are used in the subsequent filtering process to form the reconstructed image. During experiments, the two TSPs were observed to have a median size of 176,485 and 79,046 nodes and patch ordering method was spending about 85% of the execution time in computing solutions of TSPs. The TSP problem can become very computationally expensive for large set of points [8]. Computing solutions of large TSPs in $\mathbb{R}^n$ is the most computationally demanding step of the patch ordering method [6].

A modified patch ordering method for image denoising is proposed in which several smaller TSPs are formed instead of two large TSPs. The proposed method's denoising performance is comparable to the original method. Overall, the proposed method can perform denoising in 40% less time.

In Section II the original patch ordering method's formulation for denoising are recapped. In Section III the proposed modified patch ordering method is described. In Section IV the experimental setup is described. In Section V performance of the proposed method is discussed.

## II. PATCH ORDERING FOR IMAGE DENOISING

Using $\sqrt{n}$ by $\sqrt{n}$ overlapping image patches, the patch ordering method for image processing is given by

$$\hat{\mathbf{y}} = \mathbf{D}^{-1} \sum_{j=1}^{n} \mathbf{R}_j^T \mathbf{P}^{-1} H(\mathbf{P}\mathbf{R}_j \mathbf{z}) \qquad (1)$$

To denoise an image corrupted by Gaussian noise, all overlapping patches of the noisy image $\mathbf{z}$ (in column stacked form) are divided into two categories. Based on the standard deviation of its pixel values a patch is categorized as smooth or edge and texture [6]. Let the set *Ss* contain all smooth patches and the set *Se* contain all patches with edge and texture. Next, every overlapping sub-image, $\tilde{\mathbf{z}}_j = \mathbf{R}_j \mathbf{z}$, is divided into two signals $\tilde{\mathbf{z}}_{j,s}$ and $\tilde{\mathbf{z}}_{j,e}$. The signal $\tilde{\mathbf{z}}_{j,s}$ is made of $j$th sub-image's pixels corresponding to smooth patches and the signal $\tilde{\mathbf{z}}_{j,e}$ is made of $j$th sub-image's pixels corresponding to edge and texture patches. A matrix $\mathbf{P}_s$ is constructed that extracts $\tilde{\mathbf{z}}_{j,s}$ from $\tilde{\mathbf{z}}_j$, and applies a permutation to it. Also, matrix $\mathbf{P}_e$ is constructed to extract $\tilde{\mathbf{z}}_{j,e}$ from $\tilde{\mathbf{z}}_j$, and permute it. To obtain $\tilde{\mathbf{z}}_j^p = \mathbf{P}\tilde{\mathbf{z}}_j$, the $j$th sub-image with ordered pixels, $\tilde{\mathbf{z}}_j^p$, and the permutation matrix $\mathbf{P}$ are defined as

$$\tilde{\mathbf{z}}_j^p = \begin{bmatrix} \tilde{\mathbf{z}}_{j,s}^p \\ \tilde{\mathbf{z}}_{j,e}^p \end{bmatrix}, \qquad \mathbf{P} = \begin{bmatrix} \mathbf{P}_s \\ \mathbf{P}_e \end{bmatrix}. \qquad (2)$$

The matrix $\mathbf{P}_s$ is constructed using solution of the TSP formulated by considering patches in set *Ss* as points in $\mathbb{R}^n$. Similarly, permutation matrix $\mathbf{P}_e$ is constructed using solution of the TSP formulated by treating all patches in set *Se* as points in $\mathbb{R}^n$. $n$ is the number of pixels per patch. Depending on noise intensity $n$ ranges from 25 to 64 with larger values for processing images with higher noise [6].

It has been demonstrated that use of multiple permutation matrices, **P,** gives improved results [6]. When K permutation matrices are employed in patch ordering, the reconstructed

Badre Munir is with the Faculty of Computer Science and Engineering, GIK Institute of Engineering Sciences and Technology, Topi, Pakistan. (e-mail: badr@giki.edu.pk)

image $\hat{\mathbf{y}}$ is given by

$$\hat{\mathbf{y}} = \frac{1}{K}\sum_{k=1}^{K}\left(\mathbf{D}^{-1}\sum_{j=1}^{n}\mathbf{R}_j^T\mathbf{P}_k^{-1}H\mathbf{P}_k\mathbf{R}_j\mathbf{z}\right). \quad (3)$$

The 1D smoothing operator $H$ corresponds to use of two pre-learned filters $\mathbf{h}_s$ and $\mathbf{h}_e$ applied on the ordered signals $\tilde{\mathbf{z}}_{j,s}^p$ and $\tilde{\mathbf{z}}_{j,e}^p$ respectively. Steps for learning the 25-tap filters $\mathbf{h}_s$ and $\mathbf{h}_e$ can be found in [6]. With convolution matrices for ordered signals $\tilde{\mathbf{z}}_{j,s}^p$ and $\tilde{\mathbf{z}}_{j,e}^p$ denoted by $\mathbf{M}_{j,s}^p$ and $\mathbf{M}_{j,e}^p$ respectively, the equation for image denoising via patch ordering method takes the form

$$\hat{\mathbf{y}} = \frac{1}{K}\sum_{k=1}^{K}\left(\mathbf{D}^{-1}\sum_{j=1}^{n}\mathbf{R}_j^T\mathbf{P}_k^{-1}\mathbf{M}_j^p\mathbf{h}\right) \quad (4)$$

with $\mathbf{M}_j^p$ and $\mathbf{h}$ given by

$$\mathbf{M}_j^p = \begin{bmatrix}\mathbf{M}_{j,s}^p & 0 \\ 0 & \mathbf{M}_{j,e}^p\end{bmatrix}, \quad \mathbf{h} = \begin{bmatrix}\mathbf{h}_s \\ \mathbf{h}_e\end{bmatrix}. \quad (5)$$

It has been shown in [6] that K=10 is a good compromise for image denoising. Therefore, ten permutation matrices $\mathbf{P}_k$ are constructed. Constructing one permutation matrix requires solutions of two large TSPs. With K=10, this corresponds to computing twenty TSP solutions for denoising an image. It is well known that solving large TSPs is computationally intensive [8]. Computing ten solutions of large TSPs is the most computationally intensive step of patch ordering method [6]. During experiments, this step was observed to consume a substantial 85% of patch ordering method's total execution time.

### III. Proposed Approach

To speed up patch ordering method for image denoising, changes are proposed that reduce the time required for constructing ten permutation matrices. Like the original patch ordering method, all overlapping patches are first divided into the two sets $Ss$ and $Se$. But now, partitions of both sets $Ss$ and $Se$ are formed. Specifically, subsets $Ss_i$ and $Se_j$ are formed such that

$$Ss = \bigcup_{i=1}^{W}Ss_i, \quad Se = \bigcup_{j=1}^{X}Se_j. \quad (6)$$

The image patches are now contained in a total of W+X disjoint sets. A consequence of forming these subsets is that now every overlapping sub-image, $\tilde{\mathbf{z}}_j = \mathbf{R}_j\mathbf{z}$, is divided into W+X shorter signals. All the W+X component signals of $j$th overlapping sub-image, $\tilde{\mathbf{z}}_j$, are permuted by different permutation matrices. For example, the component signal $\tilde{\mathbf{z}}_{j,s2}$ is made of $j$th sub-image's pixels belonging to patches contained in second subset of set $Ss$. A matrix $\mathbf{P}_{s2}$ is constructed that extracts $\tilde{\mathbf{z}}_{j,s2}$ from $\tilde{\mathbf{z}}_j$ and applies a permutation to it yielding ordered signal $\tilde{\mathbf{z}}_{j,s2}^p$. As another example, the signal $\tilde{\mathbf{z}}_{j,e5}$ is made of $j$th sub-image's pixels belonging to patches contained in fifth subset of set $Se$. A matrix $\mathbf{P}_{e5}$ extracts $\tilde{\mathbf{z}}_{j,e5}$ from $\tilde{\mathbf{z}}_j$ and applies a permutation to it yielding ordered signal $\tilde{\mathbf{z}}_{j,e5}^p$. To obtain $\tilde{\mathbf{z}}_j^p = \mathbf{P}\tilde{\mathbf{z}}_j$, the $j$th sub-image with piecewise ordered pixels, $\tilde{\mathbf{z}}_j^p$, and permutation matrix $\mathbf{P}$ are defined in proposed method as

$$\tilde{\mathbf{z}}_j^p = \begin{bmatrix}\tilde{\mathbf{z}}_{j,s1}^p \\ \tilde{\mathbf{z}}_{j,s2}^p \\ \vdots \\ \tilde{\mathbf{z}}_{j,sW}^p \\ \tilde{\mathbf{z}}_{j,e1}^p \\ \tilde{\mathbf{z}}_{j,e2}^p \\ \vdots \\ \tilde{\mathbf{z}}_{j,eX}^p\end{bmatrix}, \quad \mathbf{P} = \begin{bmatrix}\mathbf{P}_{s1} \\ \mathbf{P}_{s2} \\ \mathbf{P}_{s3} \\ \vdots \\ \mathbf{P}_{sW} \\ \mathbf{P}_{e1} \\ \mathbf{P}_{e2} \\ \mathbf{P}_{e3} \\ \vdots \\ \mathbf{P}_{eX}\end{bmatrix}. \quad (7)$$

To form permutation matrix, $\mathbf{P}$, patches in the W+X disjoint subsets are taken as points in $\mathbb{R}^n$ to form W+X TSPs. Solutions of these smaller TSPs are used to construct W+X permutation matrices that are combined to form permutation matrix $\mathbf{P}$. The algorithm of original patch ordering method is used to solve all W+X TSPs.

All the W+X ordered component signals of $\tilde{\mathbf{z}}_j^p$ are filtered by separate pre-learned 25-tap filters. Each filter works on one component signal of $\tilde{\mathbf{z}}_j^p$. The steps to learn filters are the same as in original patch ordering method [6]. In the proposed method, the vector $\mathbf{h}$ that stores the taps for W+X filters is given by

$$\mathbf{h} = \begin{bmatrix}\mathbf{h}_{s1} \\ \mathbf{h}_{s2} \\ \vdots \\ \mathbf{h}_{sW} \\ \mathbf{h}_{e1} \\ \mathbf{h}_{e2} \\ \vdots \\ \mathbf{h}_{eX}\end{bmatrix}. \quad (8)$$

In the proposed method, the convolution matrix to perform filtering on $j$th overlapping sub-image is given in (9). As examples, $\mathbf{M}_{j,s2}^p$ and $\mathbf{M}_{j,e5}^p$ denote convolution matrices for the ordered component signals $\tilde{\mathbf{z}}_{j,s2}^p$ and $\tilde{\mathbf{z}}_{j,e5}^p$ respectively. The $n$ matrices $\mathbf{M}_j^p$, $j = 1,2,...,n$ and the matrix $\mathbf{P}$ as well as the vector $\mathbf{h}$ computed by the proposed approach are used in (4) to perform image denoising.

$$\mathbf{M}_j^p = \begin{bmatrix} \mathbf{M}_{j,s1}^p & 0 & 0 & 0 & 0 & 0 & 0 & 0 \\ 0 & \mathbf{M}_{j,s2}^p & 0 & 0 & 0 & 0 & 0 & 0 \\ 0 & 0 & \ddots & 0 & 0 & 0 & 0 & 0 \\ 0 & 0 & 0 & \mathbf{M}_{j,sW}^p & 0 & 0 & 0 & 0 \\ 0 & 0 & 0 & 0 & \mathbf{M}_{j,e1}^p & 0 & 0 & 0 \\ 0 & 0 & 0 & 0 & 0 & \mathbf{M}_{j,e2}^p & 0 & 0 \\ 0 & 0 & 0 & 0 & 0 & 0 & \ddots & 0 \\ 0 & 0 & 0 & 0 & 0 & 0 & 0 & \mathbf{M}_{j,eX}^p \end{bmatrix} \quad (9)$$

## IV. EXPERIMENTAL SETUP

The original method and the proposed method were both executed with same experimental parameter values as used by authors in [6]. The parameter values vary with standard deviation σ of Gaussian noise added to test images. The images Hill, Man, and Couple have been used for learning the filters in [6] and they were also used for learning filters in the proposed method. The images Barbara, Boats, and Lena were used as test images.

In proposed method, W subsets of set *Ss* (smooth patches) and X subsets of set *Se* (edge and texture patches) are formed (7). During experiments, the proposed method was executed under two scenarios. In the first scenario, values of W and X were chosen so that subsets of both *Ss* and *Se* contain at most 20,000 (20K) patches. As the TSPs in the proposed method are formed from subsets, sizes of TSPs were also capped at 20,000 in the first scenario. In the second scenario, the size of all subsets, and hence TSPs as well, were capped at 10,000 (10K) patches. The values of W and X for implementing the two scenarios are given in Table I.

## V. RESULTS

In the original patch ordering method for image denoising, computing ten solutions of two large TSPs has been reported as the most computationally intensive step [6]. During experiments, this step was observed to consume 85% of total execution time. Results show that the proposed method requires less time for solving TSPs and gives denoising performance similar to the original method.

Fig. 1 shows the time spent in computing ten solutions of TSPs in both the original and the proposed method. In the original method ten solutions of two large TSPs are computed and in the proposed method solutions of W+X smaller TSPs are computed. The proposed method was executed under two scenarios and times for both scenarios are shown in Fig. 1. In the first scenario, the proposed method took 30% less time for computing ten solutions of all TSPs. In the second scenario, this time further reduced to half of time taken by original method. In the first scenario, proposed method was executed with size of TSPs capped at 20,000 patches (Prop20K) whereas in second scenario it was capped at 10,000 patches (Prop10K). The results demonstrate that the proposed method can complete the most computationally demanding step of computing TSPs solution in half as much time taken by original method. Both methods were executed over an Intel® Core™ i3-2120 CPU.

In both original and proposed method, all overlapping sub-images are filtered and used in forming the reconstructed image. However, the two methods filter sub-images differently. To filter a sub-image, the original method forms two lengthy signals whereas the proposed method forms W+X shorter signals. In the original method, pixels of every sub-image are divided into two disjoint sets yielding two 1D signals. In the proposed method, however, pixels of every sub-image are divided into W+X disjoint sets to form W+X 1D signals. In both methods, separate pre-learned filters smooth the 1D constituent signals to form a filtered sub-image.

Table II lists PSNR values of the three test images denoised by the original method as well as the proposed method. The latter is executed under two scenarios. In the first scenario, the proposed method's performance differed by 0.032 dB on average. In the second scenario, the average difference reduced to 0.016 dB. In the first scenario, length of constituent signals was capped at 20,000 (Prop20K). In the second scenario, the cap was at 10,000 (Prop10K). The results demonstrate that the proposed method can yield denoising results extremely close to the original patch ordering method despite the difference in filtering of sub-images.

## VI. CONCLUSION

Smooth ordering of local patches (patch ordering) has been shown to give state-of-art results for image denoising. However, using solutions of large TSPs makes it a computationally intensive method and about 85% of execution time is consumed in computing TSPs' solutions. In the proposed method, two changes are made to the original patch ordering method. First, several smaller TSPs are formed instead of two large TSPs. Second, the filtering process is varied to work with solutions of several smaller TSPs. These changes can halve the time spent on computing TSPs' solutions. At the same time, the proposed method performs extremely close to the original method in terms of PSNR. Overall, the proposed method can denoise images in about 40% less time. The proposed approach may aid in speeding up patch ordering method for other image processing tasks.

TABLE I
VALUES OF PARAMETERS W AND X USED IN EXPERIMENTS FOR EXECUTING THE PROPOSED METHOD. TO DENOISE AN IMAGE CORRUPTED WITH GAUSSIAN NOISE OF STANDARD DEVIATION σ, THE PROPOSED METHOD WAS EXECUTED UNDER TWO SCENARIOS:
1) THE SIZE OF SUBSETS AND TSPS CAPPED AT 20,000 (PROP20K).
2) THE SIZE OF SUBSETS AND TSPS CAPPED AT 10,000 (PROP10K).

| σ | Prop20K | | Prop10K | |
|---|---|---|---|---|
| | W | X | W | X |
| 25 | 10 | 6 | 19 | 12 |
| 50 | 11 | 5 | 21 | 9 |

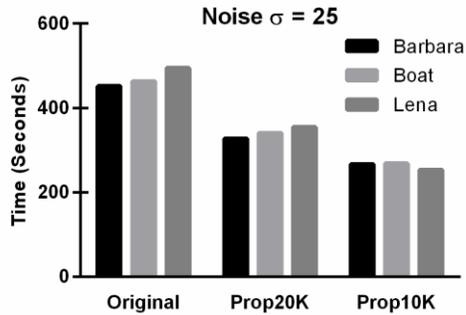

(a)

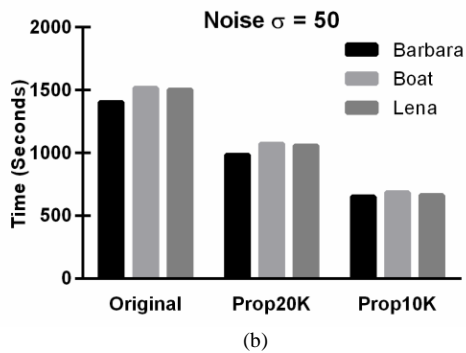

(b)

Fig. 1. For image denoising, time spent on computing solutions of TSPs in original patch ordering method (Original) and in the proposed method run under two scenarios. The proposed method was executed with TSPs' size limited to 20,000 (Prop20K) and 10,000 (Prop10K). Images were corrupted with Gaussian noise of standard deviation (a) σ = 25 and (b) σ = 50.

TABLE II
DENOISING RESULTS (PSNR IN dB) OF NOISY VERSIONS OF 3 IMAGES, OBTAINED WITH THE ORIGINAL PATCH ORDERING METHOD AND THE PROPOSED METHOD EXECUTED UNDER TWO SCENARIOS WITH TSPS' SIZE LIMITED TO 20,000 (PROP20K) AND 10,000 (PROP10K).

| Image | Method | σ /PSNR | |
|---|---|---|---|
| | | 25/20.25 | 50/14.64 |
| Barbara | Original | 30.36 | 26.97 |
| | Prop20K | 30.35 | 26.93 |
| | Prop10K | 30.30 | 26.15 |
| Boat | Original | 29.50 | 26.15 |
| | Prop20K | 29.46 | 26.15 |
| | Prop10K | 29.44 | 26.15 |
| Lena | Original | 31.54 | 28.47 |
| | Prop20K | 31.54 | 28.44 |
| | Prop10K | 31.54 | 28.48 |